\documentclass{article}
\usepackage{spconf,amsmath,epsfig}
\usepackage{graphicx}
\usepackage{booktabs}
\usepackage{pifont}
\usepackage{color}

\let\OLDthebibliography\thebibliography
\renewcommand\thebibliography[1]{
  \OLDthebibliography{#1}
  \setlength{\parskip}{0pt}
  \setlength{\itemsep}{0pt plus 0.3ex}
}

\begin{document}\sloppy

\def\x{{\mathbf x}}
\def\L{{\cal L}}

\title{TINA: Think, Interaction, and Action Framework for Zero-Shot Vision Language Navigation}
%
\name{Dingbang Li$^{1}$ \qquad Wenzhou Chen$^{2}$ \qquad Xin Lin$^{1, \star}$}

\address{$^{1}$ East China Normal University \\ 
$^{2}$ Hangzhou Dianzi University
}

\maketitle

%

\begin{abstract}
Zero-shot navigation is a critical challenge in Vision-Language Navigation (VLN) tasks, where the ability to adapt to unfamiliar instructions and to act in unknown environments is essential. Existing supervised learning-based models, trained using annotated data through reinforcement learning, exhibit limitations in generalization capabilities. Large Language Models (LLMs), with their extensive knowledge and emergent reasoning abilities, present a potential pathway for achieving zero-shot navigation. This paper presents a VLN agent based on LLMs, exploring approaches to the zero-shot navigation problem. To compensate for the shortcomings of LLMs in environmental perception, we propose the Thinking, Interacting, and Action (TINA) framework. TINA enables the agent to scrutinize perceptual information and autonomously query key clues within the environment through an introduced question-answering module, thereby aligning instructions with specific perceptual data. The navigation agent's perceptual abilities are enhanced through the TINA framework, while the explicit thought and query processes also improve the navigational procedure's explainability and transparency. We evaluate the performance of our method on the Room-to-Room dataset. The experiment results indicate that our approach improves the navigation performance of LLM-based agents. Our approach also outperformed some supervised learning-based methods, highlighting its efficacy in zero-shot navigation.
\end{abstract}
\section{Introduction}
\label{sec:intro}

\begin{figure}[htb]
\centering
\centerline{\includegraphics[width=\linewidth]{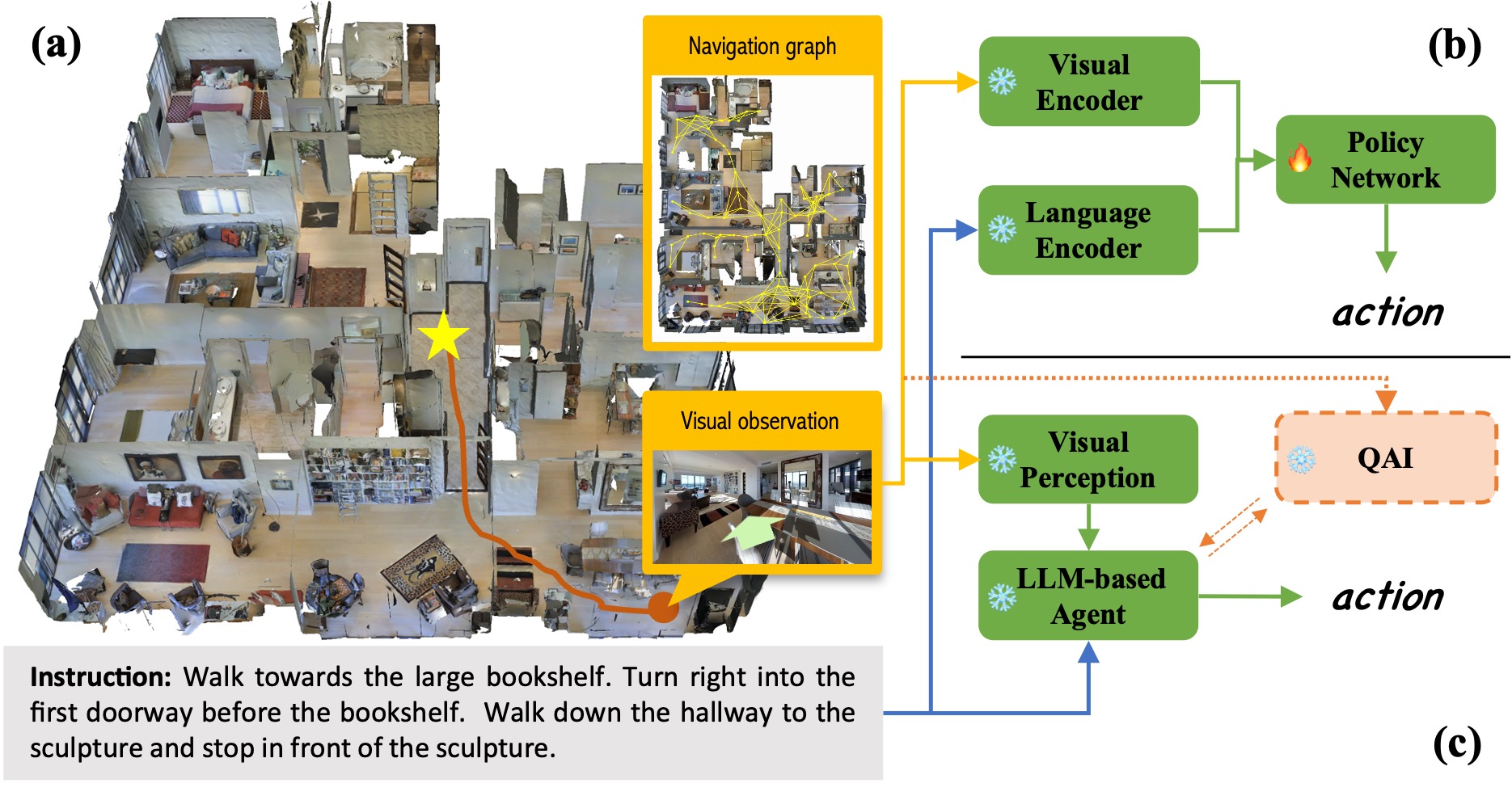}}
\caption{\textbf{(a)} A VLN Examples. \textbf{(b)} Supervised learning methods train a policy network based on pre-trained visual encoders and text embeddings. \textbf{(c)} LLM-based agents utilize an LLMs for reasoning. Our framework introduces additional modules to enhance the agent's capabilities. The snowflake symbol indicates frozen parameters, the flame signifies trainable ones.}
\label{fig:vln}
\end{figure}

The advancement of computer vision and natural language processing has facilitated research in vision-language fusion, such as Visual Question Answering\cite{antol2015vqa} (VQA), Image Caption\cite{xu2015show}, and Vision-Language Navigation\cite{anderson2018vision} (VLN). In VLN tasks, agents must navigate through diverse environments based on natural language instructions, necessitating multifaceted expertise in linguistic semantics, visual perception, and dynamic decision-making (Figure \ref{fig:vln}). While supervised deep learning has propelled the development of VLN models, existing models exhibit limitations in generalization and decision-making transparency\cite{hong2021vln, lin2022adapt, fried2018speaker, tan2019learning, anderson2019chasing}. They still lack the zero-shot ability\cite{romera2015embarrassingly} to interpret unfamiliar instructions and navigate unseen environments,  which limits the development of these models for broader applications.

Large Language Models (LLMs) have recently received considerable attention due to their remarkable language generation capabilities and extensive knowledge\cite{touvron2302llama,wei2022chain}. Research indicates that as LLMs scale, they demonstrate emergent abilities that expand their applicability across various domains, including reasoning and decision-making tasks\cite{wei2022emergent}. Recent research has revealed that, despite being trained, current VLN models still encounter difficulties when dealing with diverse instructions\cite{zhang2023vln}. For the VLN task, LLMs hold promise for enabling the zero-shot capability\cite{kojima2022large,chen20232,liang2023mo}. They can leverage their extensive knowledge and common sense reasoning to interpret and decompose new instructions and reason according to the environment, thus enabling navigation agents to adapt to unseen environments without prior training. However, a challenge arises in developing LLM-based VLN systems due to LLMs' limited visual perceptual capabilities\cite{berrios2023towards}, primarily trained on textual data. Some efforts introduced visual perceptual modules, converting visual information into textual descriptions for analysis by LLMs\cite{liu2023internchat, zhu2023minigpt,zhou2023navgpt}. While these approaches give LLM-based agents a rudimentary understanding of visual content, descriptions often remain generalized and lack specificity. In cases where visual perception needs to align with specific instructions, these visual descriptions might omit crucial information.

\begin{figure*}[htb]
\centering
\centerline{\includegraphics[width=\linewidth]{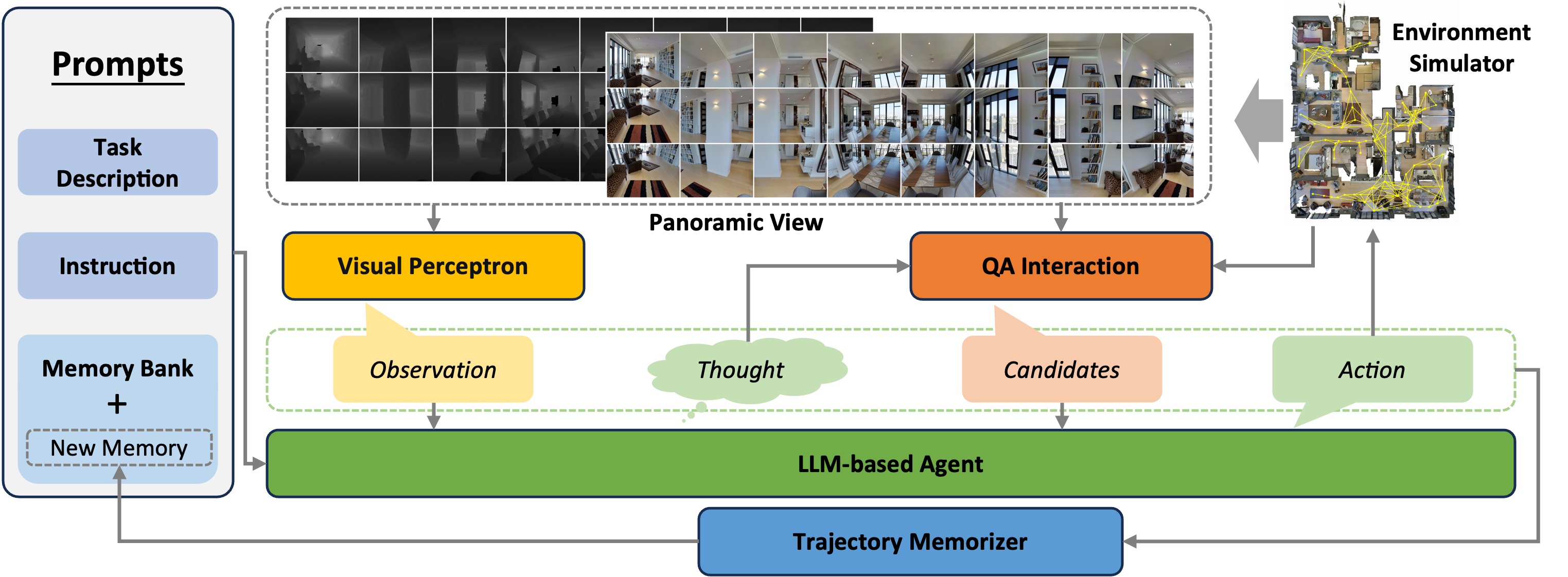}}
\caption{The schematic diagram of the TINA framework. It primarily consists of the core LLM-based agent and three peripheral modules. 1) The \textbf{Visual Perception module} is used to acquire descriptions of the surrounding environment and distance information related to various objects. 2) The \textbf{Question-Answering Interaction module} performs targeted clue queries on Candidates' visual images based on the agent's reasoning Thoughts. 3) The \textbf{Trajectory Memorizer} summarizes observations and actions from this step, storing new memories in the memory bank.}
\label{fig:agent}
\end{figure*}


To enhance the perceptual capabilities of LLMs for specific targets, we propose the TINA framework — Think, Interaction, and Action — which endows agents with the ability to scrutinize perceptual outcomes and autonomously query specific clues. Our framework comprises the core LLM agent and three additional modules: the Visual Perception (VP) module, the Question-Answer Interaction (QAI) module, and the Trajectory Memorizer (TM) module. The VP module generates rough environmental descriptions, prompting the agent to reason based on these descriptions in response to specific perceptual demands in instructions. The framework enables the agent to review the outputs of the VP and make targeted queries through the QAI module to supplement missing perceptual information. TINA extends the agent's perception through the QAI module, aligning instructions with the environment. Additionally, our framework includes a memory bank that stores the agent's actions in each round, enhancing its dynamic adaptation capabilities while filtering out redundant historical information. In the subsequent sections, we will provide a detailed description of our TINA framework for zero-shot VLN, along with insights from our experimental results and findings.

\section{Method}
\label{sec:approach}

\subsection{Preliminaries}
For a specific environment, there exists a navigation graph $G=(V,E)$, comprised of a viewpoint set $V$ and a path set $E$. The VLN task requires the agent to move from an initial viewpoint to a designated target viewpoint based on a navigation instruction $I$. The agent's action at each step is to choose the next viewpoint. The viewpoints visited during navigation are sequentially constructed into a trajectory $R = (v_0, v_1, ...,v_T)$, also representing the sequence of agent actions. When the agent arrives at a specific viewpoint $v_t$, it can access the panoramic observation $O_t$ and a navigable viewpoint set $C_t$. The panoramic observation $O_t= \{(o_{t,p}, h_{t,p}^{r}, e_{t,p}^{r})\}$ encompasses local information obtained from diverse directions, where $h^r$ and $e^r$ represent the heading angle and elevation angle relative to the agent, $o_{t,p}$ denotes the local view in that direction. Set $C_t$ encompasses adjacent viewpoints $\{(v_q, o^c_q, h_q^r, e_q^r)|v_q \in \mathcal{N}(v_t|G)\}$, $o^c_q$ is the local view in the direction of $v_q$. At step $t$, the agent selects a viewpoint from $C_t$ as the action $v_t$, and then adjusts its orientation and position to update its status from $s_t$ to $s_{t+1} = (v_{t+1}, h_{t+1}, e_{t+1})$, where $h$ and $e$ indicate the orientation of the agent itself. The agent can be regarded as a policy function for computing the probability $P(v_t|s_t,O_t,C_t,R_{[:t-1]} I;\Theta)$, wherein $\Theta$ is the function's parameters. In supervised learning, the parameters $\Theta$ are optimized by calculating the loss between the ground truth and predicted trajectory. This study uses a pre-trained LLM with frozen parameters to explore the agent's performance in a zero-shot setting. Figure \ref{fig:vln} illustrates these two approaches' differences.

\subsection{Framework Overview}
We propose an innovative agent framework for VLN agents, comprising a  core LLM and three auxiliary modules: Visual Perception ($\mathrm{VP}$), Question-Answering Interaction ($\mathrm{QAI}$), and Trajectory Memorizer ($\mathrm{TM}$). The LLM-based agent is the central component of our framework, responsible for high-level reasoning and predicting the final actions. 

At each step $t$ in the trajectory, the agent can obtain panoramic visual images from the environment. The Visual Perception module transforms the incoming panoramic visual data into initial textual descriptions denoted as $d_t$. The LLM-based agent conducts reasoning based on global instruction and the visual description $d_t$, generating intermediate reasoning results represented as Thought $h_t$. The core of this framework lies in the Question-Answering Interaction module, which plays a critical role in assessing the compatibility between potential candidate viewpoints and the agent's Thoughts. It queries the local views of the candidate viewpoints through a question-and-answer interaction, searching for crucial clues aligned with the reasoning. The query results are integrated into the descriptions of the candidate viewpoints, supplementing the information overlooked by the VP module. Subsequently, the agent selects one viewpoint from the enhanced set of candidate viewpoints as action $v_t$.
\begin{align}
    d_t &= \mathrm{VP}(O_t) \\
    h_t &= \mathrm{LLM}(d_t | I, M_{[:t-1]}) \\
    d^c_t &= \mathrm{QAI}(h_t, C_t) \\
    v_t &= \mathrm{LLM}(d^c_t | I, M_{[:t-1]}) 
\end{align}

To ensure the agent understands the task requirements, we provide a detailed description in the prompt, including task setup, module introductions, and input-output formats. Following the task description, we also offer a global instruction for each navigation case. As the agent moves through the environment, the execution history of its trajectory impacts its decision-making\cite{chen2021history}. Therefore, we establish a textual memory bank $M$ to record the agent's historical actions and input it into the agent as a prompt. These three components together form the preliminary prompt, which is conveyed to the agent at the beginning of each step before it receives new observations, as illustrated in Figure \ref{fig:agent}.

\subsection{Observation Snapshot}
The Visual Perception (VP) module translates visual information from the environment into textual descriptions that LLMs can understand. To enable the agent to perceive information from different directions, we ask the agent to gather visual images from 24 different directions, forming a panoramic observation (eight directions horizontally, with images captured at intervals of 45 degrees, and three directions vertically, covering upward / downward 30 degrees, and straight ahead at 0 degrees). Initially, we employ the BLIP-2 image captioning model\cite{li2023blip} to generate textual descriptions for each of the 24 directions. We have observed that in most cases, the agent primarily relies on information from the straight-ahead view for navigation. Therefore, we use another LLM to consolidate the content of the three descriptions in the vertical direction, resulting in eight distinct descriptions for navigation in different directions.
\begin{figure}[htb]
\centering
\centerline{\includegraphics[width=\linewidth]{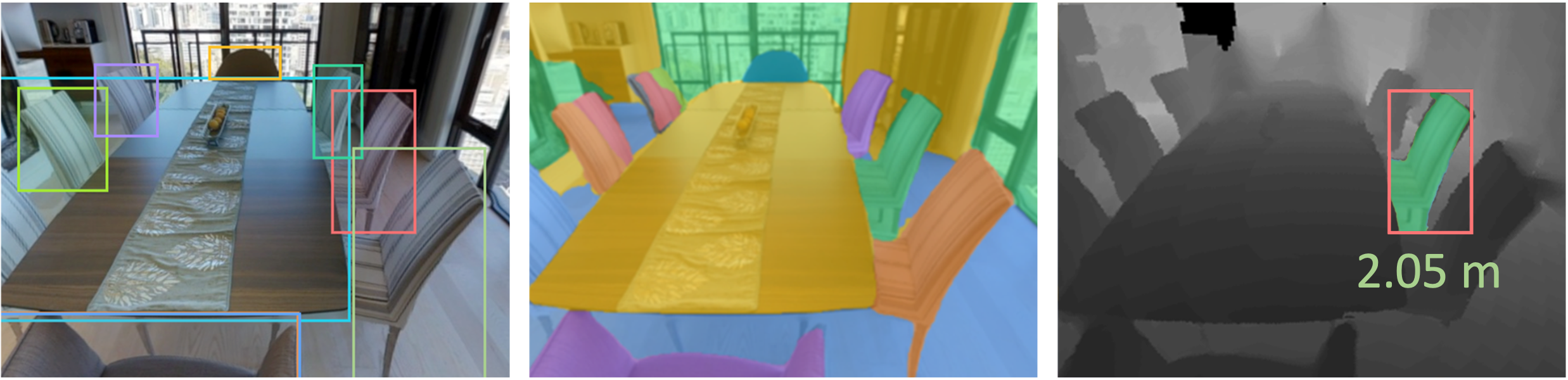}}
\caption{The figure shows how to obtain an object's distance.}
\label{fig:vp}
\end{figure}

We use IOU to match the results of DETR object detection\cite{carion2020end} and Mask2Former segmentation\cite{cheng2022masked} to enhance the agent's perception of distance to surrounding objects. We obtain the pixels of object $i$ in the RGB image by intersecting the bounding box $b_i$ and the mask $k_i$, then map them to the depth map, as shown in Figure \ref{fig:vp}. We calculate the average value of the pixels in the depth map as the distance $l_i$ between object $i$ and the agent:
\begin{align}
    l_i &= \frac{1}{|S_i|}\Sigma_{(x,y)\in S_i}D_{(x,y)} \\
    S_i &= b_i \cap k_i
\end{align}

\noindent where $D_{(x,y)}$ is the value of the coordinate $(x, y)$ in the depth map. We filter objects within a 3-meter range and incorporate their distance information into the corresponding directional descriptions.

\subsection{Interaction for Candidates Investigation}
The VP module provides the agent with environmental awareness by analyzing images from multiple discrete directions. However, this approach poses two problems for VLN. First, candidates may be located in arbitrary directions relative to the agent, leading to the segmentation of images in that direction by the VP module. To address this, we can generate descriptions again by invoking the VP module specifically for the candidate's direction. Second, the VP module provides generic descriptions and does not dynamically generate relevant information based on the agent's reasoning state, which results in incomplete environmental perception by the agent and limits the practical significance of its reasoning. This is also a challenge faced by existing LLM-based agents\cite{huang2023instruct2act}. 

\begin{figure}[htb]
\centering
\centerline{\includegraphics[width=\linewidth]{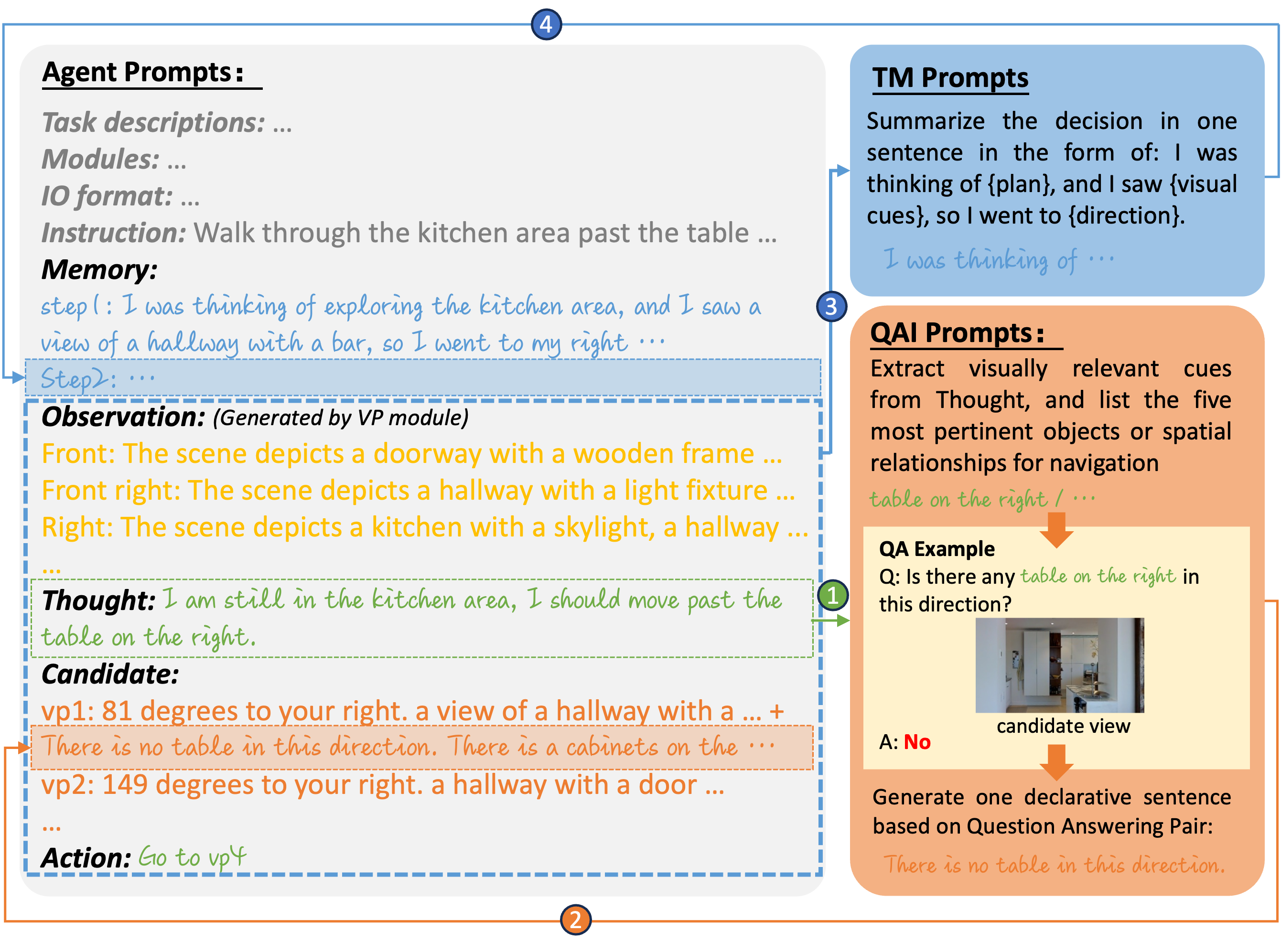}}
\caption{The key prompts and their structure for Agent navigation. The handwritten text is content generated by LLM. The numbers indicate the sequence in which the modules operate.}
\label{fig:pt}
\end{figure}

Therefore, we propose the Question Answering Interaction (QAI) module. Inspired by ReAct\cite{yao2022react}, our approach adopts a pattern of interleaved reasoning and action execution, allowing the agent to dynamically create, maintain, and adjust high-level action plans presented as Thought. This approach enhances the model's interpretability and distinguishes between the model's internal knowledge and external environmental information in scenarios involving interaction with the external environment, thus revealing the agent's need for external information during navigation. We input the agent's generated thoughts into the QAI module, which analyzes and summarizes visual cues related to these thoughts, constructing corresponding visual questions. Answers to these questions are obtained through a Visual Question Answering model from different candidate image views, as shown in Figure \ref{fig:pt}. The question-answer pairs are organized and added as supplementary content to the descriptions of the corresponding candidates. Ultimately, the agent selects an appropriate viewpoint from the optimized candidates set as current round action.

The QAI module makes full use of the agent's Thoughts, acquiring directional perceptual information and enhancing the coupling between visual perception and LLM reasoning.

\subsection{Trajectory Memorizer}
In VLN tasks, agents need to adjust their navigation strategies based on historical data dynamically. However, directly inputting all historical data into an LLM is impractical. In the later stages of navigation, the accumulated data from historical observations, reasoning, and actions can exceed the length limitations of the LLM. Additionally, a significant amount of redundant information can reduce the efficiency and accuracy of the agent's access to helpful information. To enable efficient access to the agent's historical trajectories, we introduce a memory bank $M$. After each round of reasoning, interaction, and action execution, we use a trajectory memorizer to compress and summarize the proceedings and store this new memory $m_t$ in the memory bank.  
\begin{align}
    m_t = \mathrm{TM}(d_t, h_t, d^c_t, v_t)
\end{align}

Although the memory bank grows as the trajectories accumulate, it is far smaller than the storage requirements for the entire dataset. It also filters out irrelevant information from history, retaining only key points.

\section{Experiment}
The TINA framework is implemented based on gpt-4\cite{openai2023gpt4} model. We conducted experiments on the R2R dataset\cite{anderson2018vision}, comparing the model's performance on the validation unseen split with previous methods. We also randomly sampled a portion of the entire dataset to create a dev split for analyzing the framework's modules. We adopt the following navigation metrics for evaluation: Trajectory Length (TL): average path length in meters; Navigation Error (NE): average distance in meters between the final and target location; Success Rate (SR): the ratio of paths with NE less than 3 meters; Oracle SR (OSR): SR given oracle stop policy; SR penalized by Path Length (SPL). The Results are shown in Table \ref{tab:val} and Table \ref{tab:dev}.

\begin{table}[htpb]
\fontsize{9}{10}\selectfont 
\renewcommand\arraystretch{1}
\centering
    \caption{Performance on R2R validation unseen split}
    \begin{tabular}{c|c|ccccc}
       \toprule[1pt]
        Model   & ZS & TL & NE $\downarrow$ & OSR $\uparrow$ & SR $\uparrow$ &SPL $\uparrow$   \\
        \midrule[0.5pt]
        Seq2Seq\cite{anderson2018vision} & \ding{55} & 8.39      & 7.81 & 28 & 21 &-    \\
        SF\cite{fried2018speaker} & \ding{55} & -         & 6.62 & 45 & 35 &-    \\ 
        Chasing\cite{anderson2019chasing} & \ding{55} & 9.64         & 7.20 & 44 & 35  &31    \\
        EnvDrop\cite{tan2019learning} & \ding{55} & 10.70     & 5.22 & -  & 52 &48    \\ 
        \midrule[0.5pt]
        LangNav\cite{pan2023langnav} & \ding{51} & - & 7.0 & 42 & 32 & 28 \\
        NavGPT\cite{zhou2023navgpt}  & \ding{51} & 11.45     & 6.46 & 42 & 34 &29  \\
        TINA & \ding{51}    & 11.37        & \underline{5.93} & \underline{48} & \underline{37} & \underline{33}       \\
        \bottomrule[1pt]
    \end{tabular}
    \label{tab:val}
\end{table}


Table \ref{tab:val} compares TINA with several existing methods, including supervised learning baselines and the latest zero-shot work. The "ZS" in the table indicates whether the method is zero-shot. The experimental results show that TINA has surpassed some supervised learning methods and outperformed the latest zero-shot methods, demonstrating its potential in zero-shot navigation. We also conducted several ablation experiments to investigate the roles of different components in the TINA framework. We randomly sampled a small subset of the entire dataset to create a dev split for the ablation experiments, and the results are presented in Table \ref{tab:dev}. 

First, we investigated the functionality of the QAI module. In this setting ("$w/o$ QAI"), the agent does not ask questions about the image content and relies solely on the VP's glancing descriptions for navigation. It can be observed that without the QAI module, the agent may lose crucial information, leading to a decline in navigation performance, which highlights the importance of expanding the visual perception capabilities of LLM-based agents.

\begin{table}[htpb]
\fontsize{9}{10}\selectfont 
\renewcommand\arraystretch{1}
\centering
    \caption{Ablation experiments results}
    \begin{tabular}{l|ccccc}
       \toprule[1pt]
        \multicolumn{1}{c|}{Setting} &TL & NE $\downarrow$ & OSR $\uparrow$ & SR $\uparrow$ &SPL $\uparrow$   \\
        \midrule[0.5pt]
        base                & 12.03     & \underline{7.25}  & \underline{40}    & \underline{31}    & \underline{27}       \\
        -\textit{w/o} QAI   & 12.88     & 7.81  & 37    & 29    & 26       \\
        -\textit{w/o dis} & 14.91     & 8.43  & 32    & 25    & 22       \\
        -\textit{w/o seg}   & 13.14     & 7.49  & 36    & 29    & 26       \\
        \bottomrule[1pt]
    \end{tabular}
    \label{tab:dev}
\end{table}

To explore the impact of distance perception on agent navigation, we attempted to turn off the agent's perception of distance,  shown as "$w/o~dis$". It can be seen that the absence of distance information results in a significant performance drop. When the VP stops providing distance information between objects, the scene depicted by the environment description will have greater randomness, and the agent also struggles to estimate the distance between itself and the target point. We further tested the performance gain using instance segmentation in the VP (see "$w/o~ seg$"). In this experiment, we only obtained depth information through the center points of bounding boxes, decreasing the agent's navigation success rate. We attribute this to the irregular shapes of objects, which can cause the center point of the bounding box to fall outside the object pixels, resulting in inaccurate distance calculations. Segmentation allows for a more precise alignment of RGB and depth pixels.

Through the ablation experiments, we have demonstrated that environmental perception is the most crucial factor limiting the zero-shot navigation of LLM-based agents. We propose TINA to enhance the agent's environmental perception capabilities, and the experiments indeed confirm the effectiveness of TINA. Additionally, we have found that LLM-based agents require cognitive and modeling capabilities in 3D space to adapt to a broader range of tasks. Despite some enhancements we have made to improve the agent's accuracy in obtaining distance information, the transition from 2D perception to 3D perception remains an important future research direction for LLM-based agents.

\begin{figure}[htb]
\centering
\centerline{\includegraphics[width=0.95\linewidth]{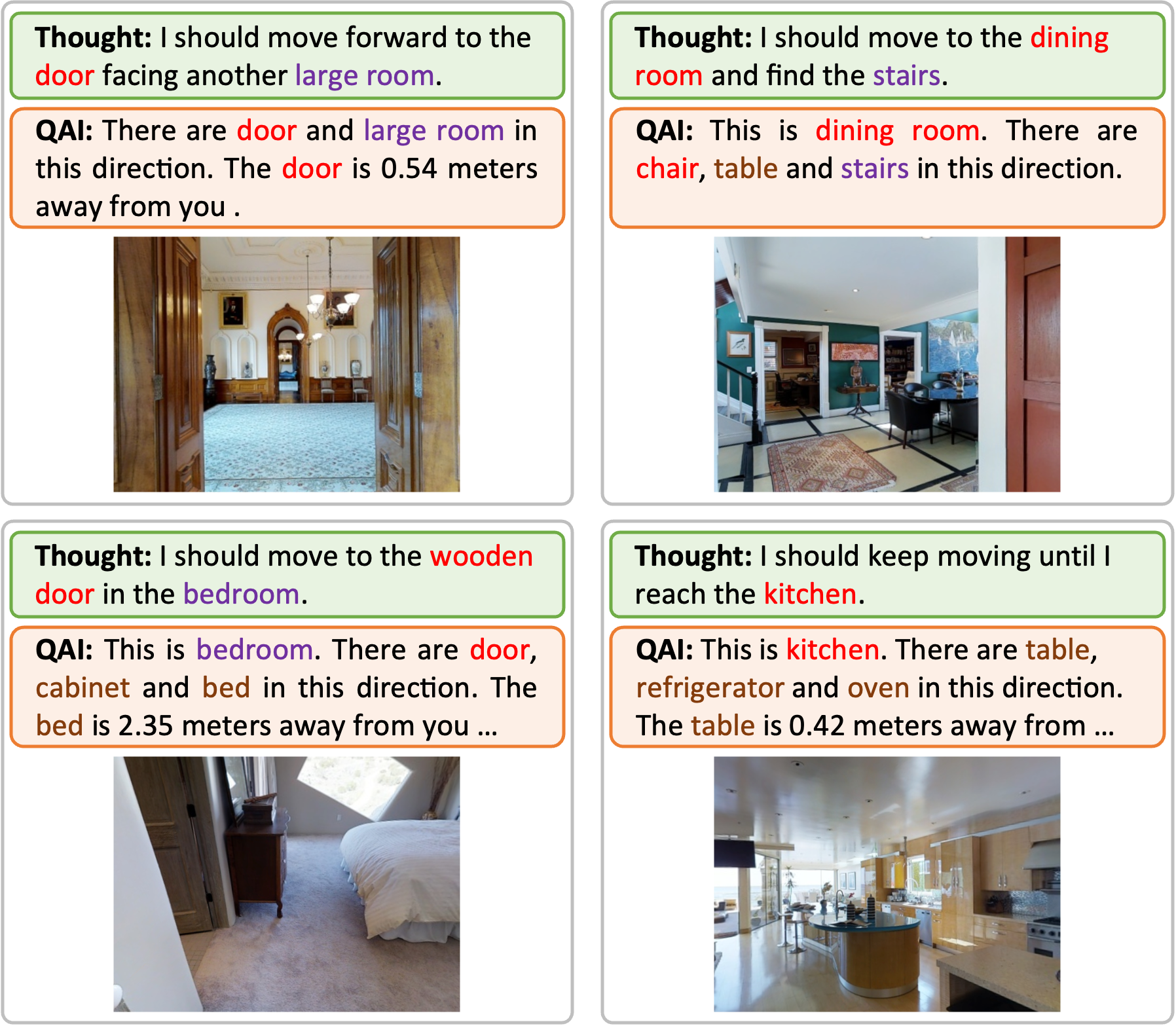}}
\caption{Some candidate viewpoints selected by the agent based on the Thought, along with the corresponding QAI output.}
\label{fig:qaivis}
\end{figure}

We also conducted a case study on the agent's and QAI's interactions, and some of the results are shown in Figure \ref{fig:qaivis}. It can be observed that the QAI module can generate descriptions closely related to Thoughts, enabling the agent to make more accurate selections for the next viewpoint, which also enhances the explainability of the navigation process.

\section{Conclusion}


In this paper, we explore the zero-shot navigation problem based on LLMs and propose the TINA framework, enabling the agent to scrutinize perceptual information and autonomously query key clues, thereby enhancing the agent's perceptual capabilities. Experimental results validate the effectiveness of our approach, which surpasses existing zero-shot navigation models and some supervised learning-based methods without the need for additional training. We also discuss the roles of each module and present examples of explainability brought about by the QAI module.

\bibliographystyle{IEEEbib}
\bibliography{icme2023template}

\end{document}